\documentclass[sigconf]{acmart}

\usepackage{multirow}
\usepackage{graphicx}
\usepackage{subcaption}
\usepackage{footmisc}

\AtBeginDocument{%
  \providecommand\BibTeX{{%
    \normalfont B\kern-0.5em{\scshape i\kern-0.25em b}\kern-0.8em\TeX}}}

\copyrightyear{2020}
\acmYear{2020}
\setcopyright{iw3c2w3}

\acmConference[WWW '20]{Proceedings of The Web Conference 2020}{April 20--24, 2020}{Taipei, Taiwan}
\acmBooktitle{Proceedings of The Web Conference 2020 (WWW '20), April 20--24, 2020, Taipei, Taiwan}
\acmPrice{}
\acmDOI{10.1145/3366423.3380066}
\acmISBN{978-1-4503-7023-3/20/04}

\begin{document}

\title{Quantifying Community Characteristics of Maternal Mortality Using Social Media}



\author{Rediet Abebe}
\authornote{Both authors contributed equally to this research.}
\email{rabebe@fas.harvard.edu}
\affiliation{%
  \institution{Harvard University}
}
\author{Salvatore Giorgi}
\authornotemark[1]
\email{sgiorgi@seas.upenn.edu}
\affiliation{%
  \institution{University of Pennsylvania}
}

\author{Anna Tedijanto}
\email{ajt232@cornell.edu}
\affiliation{%
  \institution{Cornell University}
 }

\author{Anneke Buffone}
\email{buffone.anneke@gmail.com}
\affiliation{%
  \institution{University of Pennsylvania}
 }

\author{H. Andrew Schwartz}
\email{has@cs.stonybrook.edu}
\affiliation{%
  \institution{Stony Brook University}
 }

\renewcommand{\shortauthors}{Abebe and Giorgi, et al.}
\newcommand{\sal}[1]{\textcolor{red}{$_{S}$[#1]}}
\newcommand{\andy}[1]{\textcolor{cyan}{$_{A}$[#1]}}
\newcommand{\has}[1]{\textcolor{cyan}{$_{A}$[#1]}}
\newcommand{\rediet}[1]{\textcolor{red}{$_{L}$[#1]}}
\newcommand{\anneke}[1]{\textcolor{magenta}{$_{V}$[#1]}}
\newcommand{\todo}[1]{\textcolor{blue}{[TODO: #1]} }
\begin{abstract}
While most mortality rates have decreased in the US, maternal mortality has increased and is among the highest of any OECD nation.
Extensive public health research is ongoing to better understand the characteristics of communities with relatively high or low rates. 
In this work, we explore the role that social media language can play in providing insights into such community characteristics. 
Analyzing pregnancy-related tweets generated in US counties, we reveal a diverse set of latent topics including \emph{Morning Sickness}, \emph{Celebrity Pregnancies}, and \emph{Abortion Rights}. 
We find that rates of mentioning these topics on Twitter predicts maternal mortality rates with higher accuracy than standard socioeconomic and risk variables such as income, race, and access to health-care, holding even after reducing the analysis to six topics chosen for their interpretability and connections to known risk factors. 
We then investigate psychological dimensions of community language, finding the use of less trustful, more stressed, and more negative affective language is significantly associated with higher mortality rates, while \emph{trust} and \emph{negative affect} also explain a significant portion of racial disparities in maternal mortality. 
We discuss the potential for these insights to inform actionable health interventions at the community-level.
\end{abstract}


\maketitle

\section{Introduction}  

The United States has one of the highest maternal mortality rates of any country in the Organization for Economic Cooperation and Development group ~\cite{berg2010pregnancy,measure}. 
Approximately 700 individuals die from pregnancy-related causes 
\cite{CDCpaper,CDC2,WHO} and an estimated 60\% of these deaths are suspected to be preventable \cite{CDCpaper}. 
While the international trend has seen a reduction in maternal mortality, despite increased budgets, rates in the US have more than doubled in the past 25 years \cite{WHO}.\footnote{Note, on the other hand, US infant mortality is at a historic low \cite{CDCinfant}.} 
Black and Latina mothers bear a disproportionate brunt of this burden: Black women are three to four times more likely to die during childbirth, even after controlling for numerous socioeconomic and risk factors \cite{CDCpaper}. 
These rates vary by geography: e.g., in New York City, Black women are 12 times more likely to die during childbirth than white women \cite{NYC2,NYC1}.\footnote{This issue has garnered increases attention in part due to concentrated efforts by policy-makers, advocacy groups, and celebrities, in addition to long-standing work by community organizations \cite{gillibrand,harris,warren,house,propublica}.  e.g., see collaborations between the Atlanta-based Black Mamas Matter Alliance and the Black Maternal Health Caucus.
}
Public health research has examined potential causes for maternal mortality and disparities, pointing to issues such as access to insurance, bias in health-care, segregated hospitals, and inadequate post-delivery care \cite{depression,post1,post2,site1,site2}. While it is understood that each of community, health facility and system, patient, and provider all play a part, there is an overall pervasive concern that the specific causes and mechanisms for maternal mortality and disparities are not adequately understood \cite{CDCpaper}. The WHO 
cites a ``general lack of good data -- and related analysis -- on maternal health outcomes" as a bottleneck for gaining insights into this issue~\cite{WHO}.

In this work, we seek to partially address this gap, focusing on community-level factors that characterize maternal mortality as revealed through social media language. We examine whether community variables derived from social media language data can predict community maternal mortality rates and its racial disparity. 
While emotions and language analyzed using social media data are shown to have high-efficacy in tasks ranging from predicting allergies or life satisfaction to depression or heart disease mortality
\cite{paul2011you,de2013predicting,schwartz2013characterizing,eichstaedt2015psychological,curtis2018can}, the potential of social media has yet to be examined in this manner to help shed understanding on maternal mortality at the community level. 

Our contributions in this work are in three-folds: 
\begin{itemize} 
\item We show that there is a diverse set of pregnancy-related topics ranging from \emph{Morning Sickness}, to \emph{Abortion Rights}, to \emph{Maternal Studies}. We demonstrate that these topics predict maternal mortality rates with higher accuracy than standard socioeconomic (SES), risk factors, and race. 
\item We show that a select set of six topics, chosen for their interpretability and relations to known maternal health factors, hold as much predictive power as all pregnancy-related topics. Specifically, four of these topics -- \emph{Maternal Studies, Teen Pregnancy, Abortion Rights}, and \emph{Congratulatory Remarks} -- have negative associations with mortality rates. 
\item We examine variables associated with racial disparities in maternal mortality (i.e. the difference between rates for Black women and other races), finding that language-based scores for \emph{trust} and \emph{affect} hold explanatory power for the county-level relationship between race and maternal mortality, even after controlling for standard SES and risk-factors. 
 
\end{itemize} 

\section{Background and Related Work}

\noindent \textbf{Maternal Mortality Background.} Public health research has sought better measurements of maternal mortality rates and their causes and consequences \cite{CDCpaper,measure,WHO}. There is a long line of work exploring what community, patient, hospital, provider, or systemic-level factors may contribute to high rates of mortality and disparities in the US \cite{centers2019building,howell2018reducing,louis2015racial,lu2018reducing}. At the patient-level, cardiovascular conditions, which are related to stress, cause about one third of all pregnancy-related deaths~\cite{CDCpaper}. At the community and systemic-level, studies have shown that delivery site, segregation, and discrimination in maternity care during visits all play a role \cite{site1,site2,post1,post2}. At the systemic-level, sociological and economic research have shown racial disparities in mortality and life-expectancy \cite{chetty2016association,levine2016black}.  In line with such studies, there are numerous calls to use a data-driven approach to better grasp the role and causes of maternal mortality related to each of the above main categories \cite{CDCpaper}.

\vspace{2mm} 
\noindent \textbf{Social Media Data for Health.} Twitter data and more generally social media data has been a popular source for exploring community-level health measurements \cite{dredze}. Examples include excessive alcohol consumption \cite{curtis2018can}, depression \cite{de2013social,mowery2016towards}, heart disease, \cite{eichstaedt2015psychological}, and more generally population health and well-being \cite{culotta2014estimating,gibbons2019twitter,schwartz2013characterizing}. In addition to measuring community-level insights, these data sources have been used to study health information seeking and sharing \cite{de2014seeking} and individual-level predictions \cite{de2013predicting}. In recent years, there has also been interest in understanding the societal and ethical implications and limitations around the use of social media data for health studies and roles for computing as a diagnostic of social problems \cite{abeberoles,althoff2017population,chancellor2019taxonomy,chen2019can,conway2016social}.

\vspace{2mm} 
\noindent \textbf{Maternal Health.} An emerging topic of interest has been the use of language-driven analysis to understand pregnancy and maternal experiences. For instance, \citet{horvitz} studied Twitter posts to understand changes in emotions for mothers; \citet{maria} looked at narrative paths in individuals sharing childbirth stories on an online forum. Focusing on support, \citet{vydiswaran2014user,fertility,peer} looked at how online peer support and information exchange for pregnant individuals, their caregivers, and individuals experiencing fertility issues. \citet{abebe} looked at information seeking for pregnancy and breastfeeding related to HIV. To our knowledge, ours is the first work to employ a language-driven study to understand maternal mortality in the US.

\section{Data}\label{sec:data}

We used three sets of data sets for this study, described below:  

\subsection{Twitter Data and Seed-Words} 

To generate our pregnancy data set, we started with a random 10\% sample of the entire Twitter stream collected between 2009 and 2015~\cite{daniel10pct}.  We then used this data set to build two subsets: (1) pregnancy-related tweets and (2) tweets geo-located to US counties.

\vspace{1mm}
\noindent \textbf{Pregnancy-Related Tweets.} The first data set consisted of tweets related to pregnancy and birth. Tweets were pulled from the main data set if they contained the following seed-words: \textit{pregnancy, pregnant, infant, fetus, miscarriage, prenatal, trimester, complications, pregnant, birth, childbirth, pregnancies, baby, children, pregnancy, mother, newborn, child,} as well as their plural form, hashtags such as \textit{\#pregnancy}, and capitalizations such as \textit{Pregnancy}. These seed-words were selected by examining nearest neighbors from word2vec for words related to `pregnancy' and `pregnant.' 

We then manually examined a random sample of 1,000 tweets from the data set to test for relevance to pregnancy. Tweets that were deemed off-topic, such as those containing phrases like ``miscarriage of justice" were used to generate phrases for further data cleaning. 
We also randomly sampled tweets for specific seed-words and if a substantial (i.e., more than 20\%) of the tweets were unrelated to pregnancy, all tweets were removed from the data set, reducing the seed-set. 
After these cleaning steps, we kept 74.40\% of the data set, and validated in fresh sample of 1,000 tweets that over 95\% of them are related to pregnancy. 

\vspace{1mm}
\noindent \textbf{U.S. County Tweets.} The second data set consisted of tweets geo-located to U.S. counties. 
For this we used the County Tweet Lexical Bank~\cite{giorgi2018remarkable}. 
This data set was geo-located using self-reported location information (from the user description field) and latitude / longitude coordinates~\cite{schwartz2013characterizing}. 
The data were then filtered to contain only English tweets~\cite{langid}. 
We then limited our data set to Twitter users with at least 30 posts and U.S. counties with at least 100 such users. 
The final Twitter data set consisted of 2,041 U.S. counties. 

\subsection{Mortality Rates}
The World Health Organization (WHO) defines maternal mortality as ``the death of a woman while pregnant or within 42 days of termination of pregnancy, irrespective of the duration and site of the pregnancy, from any cause related to or aggravated by the pregnancy or its management but not from accidental or incidental causes" with the Centers for Disease Control and Prevention (CDC) expanding this time period to 1 year~\cite{zahr2004maternal,creanga2017pregnancy}. 
Data for maternal mortality was collected from the CDC WONDER online database \cite{cdcwonder}. We collected rates from 2009-2017, so as to match the time-span of our Twitter sample in addition to more recent years (2016 and 2017) since these rates are on the rise~\cite{CDCpaper}. Mortality rates are listed under the following International Classification of Diseases, Tenth Revision (ICD-10) categories: O00-O07 (pregnancy with abortive outcome) and O10-O99 (other complications of pregnancy, childbirth and the puerperium). The CDC suppresses data if a county experiences less than 10 deaths in a given time period for privacy reasons. Of the 2,041 counties in our Twitter set only 197 also had mortality rates (i.e., counties experiencing 10 or more deaths). 

Since the CDC does not report age-adjusted rates for counties with low mortality numbers, we took the crude rate as reported and created our own age-adjusted rate. To do this, we built a model using median age of females (American Community Survey, 2014; 5-year estimates) and predicted maternal mortality, taking the residuals as our new ``age-adjusted maternal mortality rate.'' This age-adjusted value is used throughout the paper. 
 
\subsection{Socioeconomic Measures and Risk Factors}  

In addition to mortality, we collected additional county-level variables on \emph{socioeconomics}, \emph{risk factors}, and \textit{race}. \textit{Socioeconomics} included unemployment rate, median income, and education (percentage of people with Bachelor's degrees and High School graduate percentage). 
For \textit{risk factors}, we included insurance rates and access to health-care (the ratio of population to number primary care providers). Finally, we also explored the relationship between language and maternal mortality with respect to percentage of Black individuals in each county. 
As discussed previously, the disparity in mortality rates for Black women is large and providing evidence toward the factors at play for such a disparity is a key application for our analyses. 
Additionally, to account for overall rates of birth, all analysis included a birth rate covariate (the rate per 1,000 women, aged 15-50, with births in the past 12 months).

The birth rate, race, SES variables, and insurance rates were collected from the 2014 American Community Survey (5 year estimates), whereas the primary care providers was collected from the 2017 County Health Rankings (as reported by the Area Health Resource File/American Medical Association, 2014).
We were able to obtain these values for each of the counties which met the Twitter and mortality inclusion criteria above.

Overall, we obtained data for 197 U.S. counties and county equivalents that met each of the data requirements above and conducted our study on these counties. The full list of these counties is included in the project page.\footref{projectpage} 

\section{Topics and Theoretical Linguistic Features}

We used three sets of features that will characterize maternal mortality through language. First, we created a set of automatically-derived topics built over the pregnancy-related tweets. These topics reveal a diversity of themes in discussions around pregnancy on the platform. Next, we used a small set of theoretically-driven language features -- (\emph{affect}, \emph{depression}, \emph{stress}, and \emph{trust}) -- in order to access psychological traits of a community and their relations to maternal mortality. Finally, we use a large, general set of topics (non-pregnancy related) to identify broader language patterns.

\subsection{Pregnancy-Related Topics}
We start with our data set of over 5 million pregnancy-related tweets described in Section \ref{sec:data}. We automatically extracted \emph{topics} using Latent Dirichlet Allocation (LDA) \cite{blei2003latent}. 
LDA is a generative statistical model which assumes that each document (in our case tweet) contains a distribution of topics, which in turn, are a distribution of words.
We use the Mallet software package~\cite{mccallum2002mallet}, which estimates the latent variable of the topics using Gibbs sampling~\cite{gelfand1990sampling}.
All default Mallet settings were used, except $\alpha$, which is a prior on the expected topics per document. We set $\alpha = 2$ since tweets are shorter than the typical length of documents. The number of topics is a free parameter and we chose 50 topics.\footnote{Before running the rest of our analysis, we ran LDA using 10, 20, 50, 100, and 200 topics. We selected 50 topics based on manual inspection of coherence and interpretability of the topics.}

\begin{table}[htpb!]\centering{\begin{tabular}{p{0.1\textwidth}p{0.34\textwidth}}\toprule
Topic Label & Top Weighted Words  \\ \hline
Teen \newline Pregnancy (1.34\%) & teen, rate, rates, teenage, highest, mortality, low, states, teens, higher, number, 20, country, american, united, education, lowest, population
\\ \hline
Morning \newline Sickness (0.54\%) & morning, sickness, purpose, symptoms, lives, wanted, williamson, tv, experience, cure, bra, marianne, thinking, signs, oral, teenagers, simon 
\\ \hline
Celebrity \newline Pregnancies (1.42\%) & kim, kardashian, kayne, amber, rose, beyonce, years, west, harry, finish, swear, north, who's, kayne's, taylor, sets, louis, wiz
\\ \hline
Abortion Rights (2.15\%) & women, abortion, care, health, abortions, bill, mortality, \#prolife, rights, law, gift, support, circumstances, crisis, irrelevant, \#prochoice, forced
\\ \hline
Maternal \newline Studies (2.56\%)&
risk, defects, study, health, weight, linked, flu, cancer, early, diet, drinking, smoking, blood, safe, alcohol, diabetes, autism, acid, disease, drug \\ \hline
Congratulatory \newline Remarks (3.06\%) & congrats, congratulations, :), boy, happy, love, daughter, son, <3, wait, sister, late, healthy, cousin, xx, amazing, :d, meet, proud  \\  \bottomrule
\end{tabular}}
\caption{Sample pregnancy topics with representative words}
\label{table: example tweets}
\end{table}

We find that our data reveals a rich set of themes related to pregnancy and birth. In Table \ref{table: example tweets}, we show a sample of six topics, which are hand-selected to demonstrate the breadth of topics in the data set.\footref{projectpage} The first column provides the topic label, which were hand-generated by the authors, and the frequency with which the topic occurs in the data set.\footnote{Note, since there are 50 topics, the average value is 2\%. Furthermore, since some themes, such as celebrity pregnancy, occur in more than one topic, the overall frequency of this theme in the data set is higher than the corresponding value in this table.} The last column corresponds to the top 10 most representative words for the topic. 

These above topics show that pregnancy-related discussions on Twitter can range from personal-health disclosure such as in \emph{Morning Sickness}, to political conversations related to \emph{Abortion Rights}, and light topics such as \emph{Congratulatory Remarks}. Topics that were not included in manuscript due to length constraints include \emph{Royal Baby}, \emph{Food Cravings}, and \emph{Pregnancy Timeline}. Each of these topics shows varying levels of popularity across the counties.

\subsection{Theoretical Features}

We also explore a set of theoretically-driven language features: \emph{affect}, \emph{depression}, \emph{trust}, and \emph{stress}. 
We downloaded pre-existing models to derive county-level language features including:
\begin{itemize}
    \item \textbf{affect} -- positive and negative emotional valence trained over Facebook posts~\cite{preoctiuc2016modelling}. 
    \item \textbf{depression} -- degree of depressive personality (a facet of the big five personality test) fit over social media users' language~\cite{schwartz2014towards}. 
    \item \textbf{trust} -- degree of trustfulness (how much one tends to trust persons or entities that they do not personally know) fit over social media users' language~\cite{zamani2018predicting}.
    \item \textbf{stress} -- amount of stress fit over social media users' language and Cohen's Stress scale~\cite{cohen1997measuring,guntuku2019understanding}. 
\end{itemize}

\subsection{General Topics}
Finally, we use a larger set of LDA topics built over a more general data set. By doing this in tandem with the pregnancy-related topics, we can zoom in on pregnancy-related themes while also exploring a larger set of language correlates, which might help in characterizing communities suffering from higher or lower rates of mortality.
To this end, we downloaded a set of 2,000 topic posteriors that were automatically-derived over the MyPersonality data set~\cite{schwartz2013personality}. 
These topics have been used over a large class of problems and have been found to be robust both in terms of interpretability and predictive power~\cite{eichstaedt2015psychological,park2015automatic,preotiuc2016studying,jaidka2018cross}, so they form a point of comparison for our domain-specific topics.

\section{Methods} 

To understand the relationship between community level language and maternal mortality, we perform three types of statistical analyses: (1) \emph{prediction} --- can language be used to predict mortality rates in a cross-sectional cross validation setup? (2) \emph{differential language analysis} -- can we gain insights into communities which suffer from higher or lower maternal mortality through language? and (3) \emph{mediating language analysis} --- can language be used to understand the mechanisms through which Black communities experience increased rates of maternal mortality?
All data processing, feature extraction and statistical analysis are performed using the open source Python package DLATK~\cite{schwartz2017dlatk}.

\subsection{Prediction}
We use two types of predictive models, depending on the type of independent variables. 
All non-language variables (i.e., SES and risk factors) are modeled with an ordinary least squares (OLS) regression, whereas language features use an $\ell_2$ regularized (Ridge) regression~\cite{hoerl1970ridge}. 
In addition to regularization, we also use a feature selection pipeline in all language based models, since the number of features can be larger than the number of observations ($N$=197 counties).
The pipeline first removes all low variance features and then features that were not correlated with our outcome. 
Finally, we applied Principal Component Analysis (PCA) to further reduce the number of features.
All models are evaluated in a 10-fold cross validation setup, with the Ridge regularization parameter $\alpha$ tuned on the training set within each fold. 
Predictive accuracy is measured in terms of a single Pearson correlation between the actual values and the predicted values, whereas standard errors are calculated across all 10 folds.

\subsection{Differential Language Analysis}
Differential Language Analysis (DLA) is used to identify language characterizing maternal mortality~\cite{schwartz2013personality,kern2016gaining}. 
Here we individually regress each of our language variables (i.e., pregnancy related topics and theoretical features) using an OLS regression, adding in access to health-care, birth rates, socioeconomics and risk factors as covariates. 
We adjust for multiple comparisons by applying a Benjamini---Hochberg false discovery rate correction to the significance threshold ($p<.05$)~\cite{benjamini1995controlling}. For LDA topics we visualize topics significant correlations as word clouds.
The word clouds display the top 15 most prevalent words within a topic sized according to their posterior likelihood.

\subsection{Mediating Language Analysis}
We explore the relationship between maternal mortality and the percentage of Black individuals within a county, as expressed through the county's language. 
Language based mediation analysis has been used in the past to explore the relationship between socioeconomics and excessive drinking~\cite{curtis2018can}. 
For this analysis, we residualize the crude maternal mortality rate, as reported by the CDC, on median age of female, birth rates, all socioeconomic variables (income, education and unemployment), insurance rates and rates of primary care providers. 

For each language variable, both the pregnancy related LDA topics and theoretical language features, we consider the mediating relationship between the topic (mediator), percentage Black (independent variable) and residualized maternal mortality rates (dependent variable). 
We follow the standard three-step, Baron and Kenny approach \cite{baron1986moderator}. 
Step 1: we regress our independent ($x$) and dependent variables ($y$; path $c$) in a standard OLS regression. 
Step 2: we regress the independent variable ($x$) and mediator ($m$; path $\alpha$). 
Finally, in Step 3 we create a multi-variate model and regress both the mediator ($m$; topic) and independent variable ($x$; percentage Black) with maternal mortality ($y$; path $c'$). 
The three models are as follows:
\begin{equation}
    y = cx + \beta_1 + \epsilon_1,
    \label{eq:mediation step 1}
\end{equation}
\begin{equation}
    m = \alpha x + \beta_2 + \epsilon_2,
    \label{eq:mediation step 2}
\end{equation}
\begin{equation}
    y = c'x + + \beta m + \beta_3 + \epsilon_3.
    \label{eq:mediation step 3}
\end{equation}
The mediation effect size ($c-c'$) is taken as the reduction in the effect size between the direct relationship (i.e., percentage Black and maternal mortality) and the mediated relationship. 
To test for significance, we use a Sobel $p$~\cite{sobel1982asymptotic} and correct all $p$ values for false discoveries via a Benjamini---Hochberg procedure.

\section{Results}

We begin by looking at correlations between maternal mortality and various socioeconomics and risk factors. Table \ref{table:correlations} shows the set of correlation coefficients. These results state that the percentage of the population that is Black and unemployment rate were positively correlated with maternal mortality rate and insurance access, income, and education were negatively correlated with maternal mortality rate. Additionally, birth rates were not significantly correlated with maternal mortality. Note that, in this paper, we only consider 197 counties in the US due to constraints around Twitter and county-mapped data as discussed in Section \ref{sec:data}. While the correlation values do not exactly match correlations for all US counties, the general direction of relationship between maternal mortality rates and these SES and risk-factors was the same, with those the strongest associations -- such as percent Black -- also matching. 

\begin{table}[ht]\centering
{\begin{tabular}{lc}\toprule
 & \begin{tabular}[c]{@{}c@{}}Correlation \end{tabular} \\ \hline
\emph{Birth Rates} &  \\
\hspace{3mm}Rate per 1,000 women & .10 [-.04,.24] \\
\emph{Race} &  \\
\hspace{3mm}Black (percent) & .49 [.36,.61]*** \\
\emph{Risk Factors} &  \\
\hspace{3mm}Primary Care Providers & -.23 [-.38,-.09]** \\
\hspace{3mm}Uninsured (percent) & .27 [.12,.41]*** \\
\emph{Socioeconomics} &  \\
\hspace{3mm}Income (log median) & -.42 [-.55,-.29]*** \\
\hspace{3mm}High School or more (percent) & -.14 [-.28,.01] \\
\hspace{3mm}Bachelor's Degree (percent) & -.38 [-.52,-.23]*** \\
\hspace{3mm}Unemployment (percent) & .26 [.12,.39]*** \\
\bottomrule
\end{tabular}}
\caption{Correlations with risk factors, socioeconomics, and race. All non-birth rate correlations controlled for birth rates. Reported standardized $\beta$ with 95\% confidence intervals in square brackets; ***$p <0.001$, **$p < 0.01$, *$p <.05$, after Benjamini---Hochberg correction}
\label{table:correlations}
\end{table}

\begin{figure}[!ht]
\centering
\includegraphics[width=1\columnwidth]{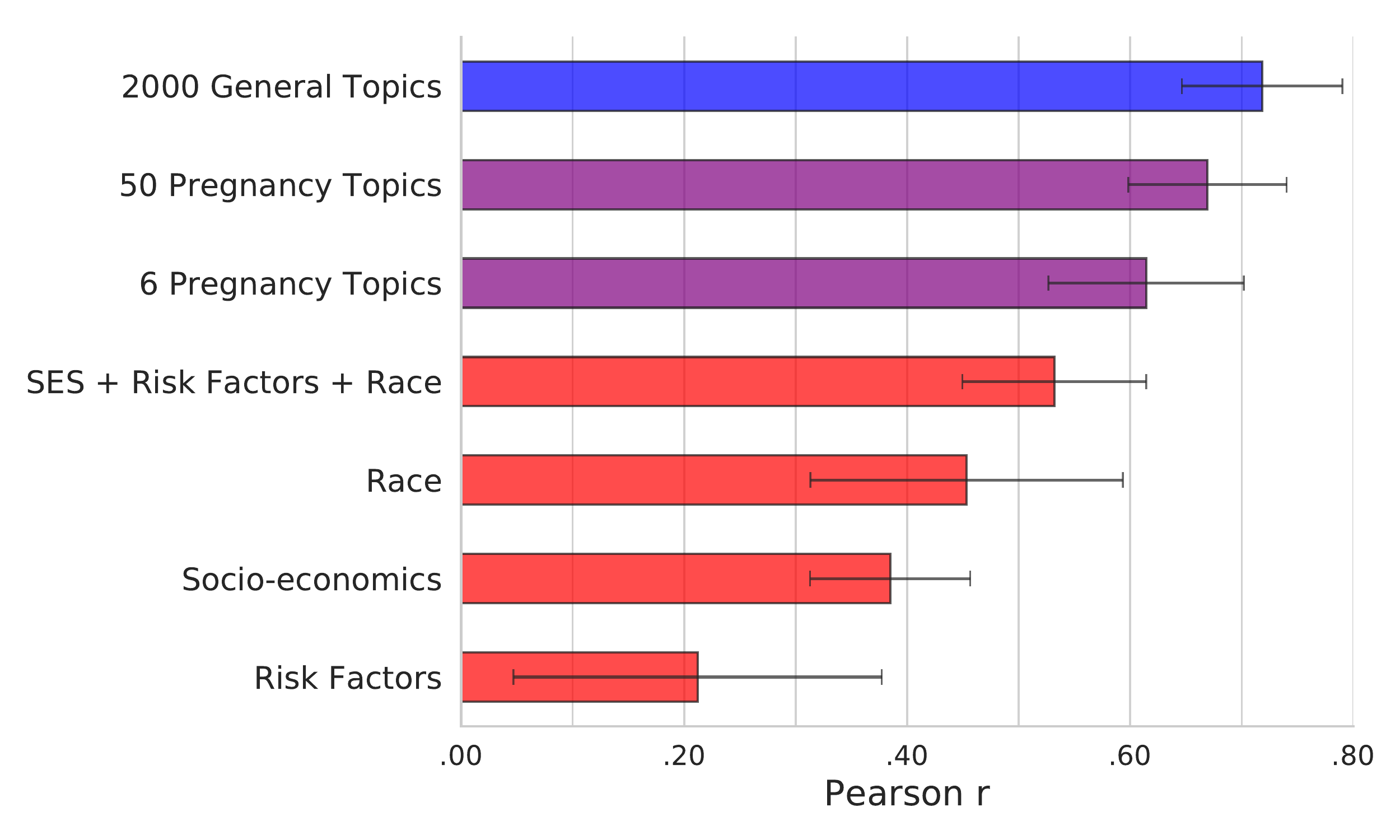}
\caption{Prediction accuracy for non-language variables (red), pregnancy related LDA topics (purple) and general set of LDA topics (blue). Reported Pearson r from 10-fold cross validation, errors bars are 95\% CI.}
\label{fig:prediction}
\end{figure}

We next look at the predictive accuracy of our 50 topics, the 2000 general topics, and the above SES and risk-factors as well as percent Black values. For this, note that we used linear regression with maternal mortality values as the outcome variable and the aforementioned language variables as the explanatory variables. Figure \ref{fig:prediction} shows that the 2000 general Facebook topics had the highest predictive power with a Pearson $r = .72$ $[.65, .79]$*** while risk factors (PCP access and insurance rate) were the lowest with a Pearson $r = .21$ $[.05, .38]$**. Overall SES factors, risk factors, and race, had significantly less predictive accuracy (using a paired t-test) than the 50 pregnancy-related topics from the Twitter data ($t=-4.63$, $p<.001$) and the 2000 general topics ($t=-4.74$, $p<.001$). 

\begin{figure}[!htbp]
{
\begin{tabular}{cccc}
  \includegraphics[height=0.29\columnwidth]{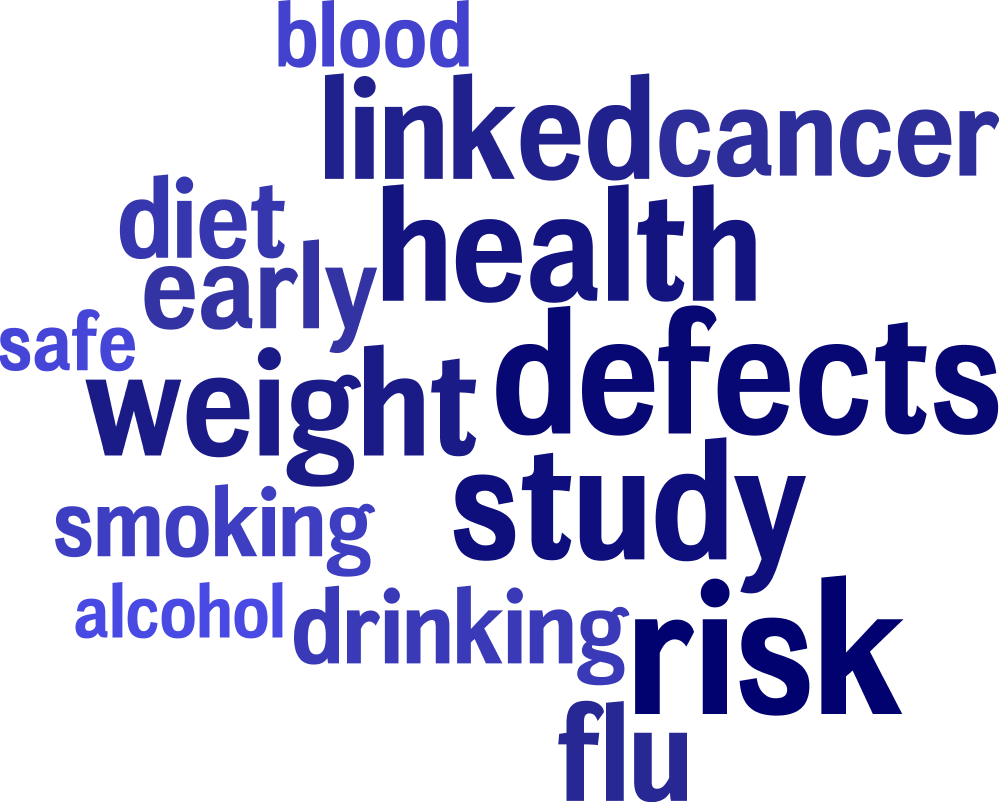} & 
  \hspace{2mm}
  \includegraphics[height=0.29\columnwidth]{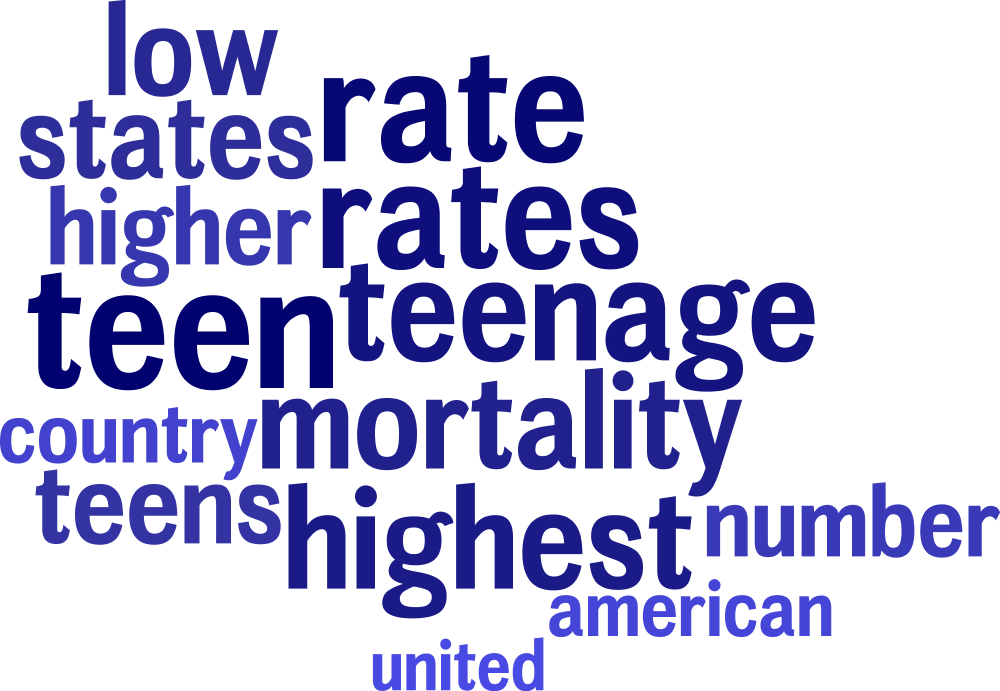}\\ 
      \hspace{1mm}
  -.38 [-.49,-.25]*** & -.35 [-.47,-.22]***\\
      \vspace{3mm}\\
  \includegraphics[height=0.24\columnwidth]{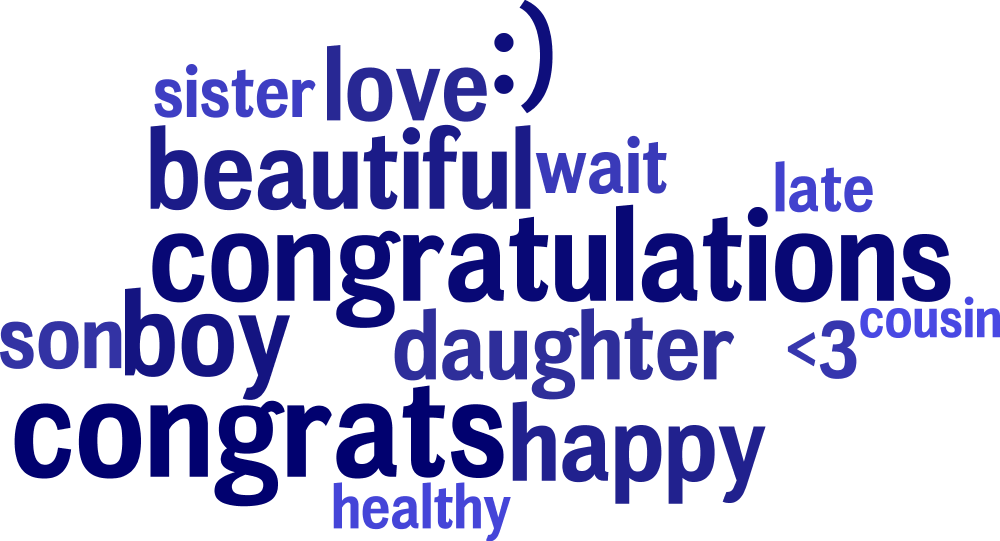}&
    \hspace{1mm}
  \includegraphics[height=0.24\columnwidth]{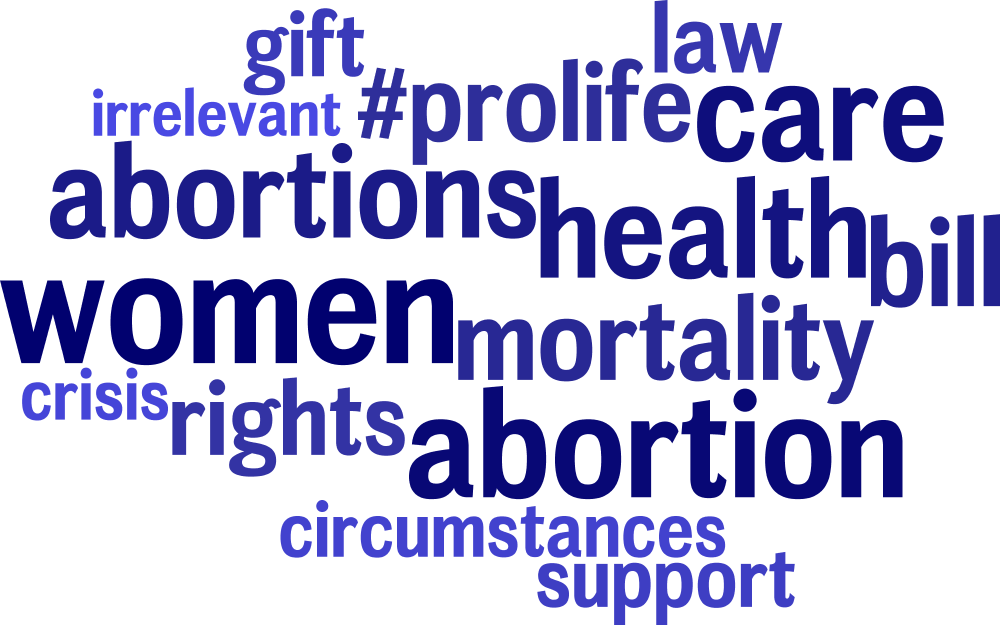} \\
    -.28 [-.40,-.14]*** & -.19 [-.32,-.05]* \\
\end{tabular}}
\caption{Differential Language Analysis using 6 pregnancy related LDA topics, controlled for race, risk factors and socioeconomics. Reported standardized $\beta$ with 95\% confidence intervals in square brackets;***$p <0.001$, **$p < 0.01$, *$p <.05$, after Benjamini---Hochberg correction.}
\label{fig:dla topics}
\end{figure}

For the Differential Language Analysis (DLA), we selected the 6 topics of interest. We ran a multi-linear regression, treating the maternal mortality rate as an outcome variable and the prevalence of these topics in the counties as the explanatory variable with birth rates, race, risk factors and socioeconomics as covariates. We found that five of the 6 topics, shown in Figure \ref{fig:dla topics} had significant associations with maternal mortality rates. \emph{Maternal studies} had the most negative association -- i.e., counties where there are relatively more tweets related to this topic had lower rates of mortality. Note that each of the four topics in the figure -- \emph{Maternal Studies, Teen Pregnancies, Congratulatory Remarks}, and \emph{Abortion Rights} -- all show negative associations with maternal mortality rates. \emph{Celebrity Pregnancies}, not shown, is positively associated ($.20$ $[.07,.33]^*$) with higher mortality.

We also used 4 theoretical features within the DLA framework: \emph{affect, depression, stress} and \emph{trust}. Results are presented in Table \ref{table:dla theoretical}. We see higher rates of maternal mortality associated with higher distrust, higher stress, higher depression, and with less affect.

\begin{table}[ht]\centering
{
\begin{tabular}{p{0.1\textwidth}p{0.15\textwidth}}\toprule
 & Correlation \\ \hline
Affect &  -.30 [-.43,-.17]*** \\
Depression & .23 [.10,.36]** \\
Stress & .24 [.10,.37]** \\
Trust &  -.38 [-.49,-.25]***  \\  \bottomrule
\end{tabular}
}
\caption{Differential Language Analysis of theoretically relevant features. Reported standardized $\beta$ with 95\% confidence intervals in square brackets;***$p <0.001$, **$p < 0.01$, *$p <.05$, after Benjamini---Hochberg correction.}
\label{table:dla theoretical}
\end{table}

Finally, we explore disparities by race at the population level. 
The county-level health disparity itself can be seen simply from the strong correlation between the two variables: communities that are more Black, have greater maternal mortality.  
We turn to Twitter-based community characteristics as mediators (i.e. explainers) of this  race-mortality relationship. 
The idea behind mediation analysis, is that if included a 3rd variable (i.e. a Twitter measurement) in the linear analysis reduces the relationship of the first 2 (i.e. race and maternal mortality), then this third variable is accounting for some of the covariance between the first two.

\begin{table}[ht]\centering
\resizebox{\columnwidth}{!}{\begin{tabular}{lccc}\toprule
 & $c-c'$ & $\alpha$ & $\beta$ \\ \hline
Affect & .11** & -.40 [-.53,-.27]*** & -.27 [-.41,-.13]*** \\
Depression & -.04 & -.26 [-.39,-.12]*** & .14 [.01,.28]* \\
Stress & -.01 & -.06 [-.20,.08] & .15 [-.02,.28]* \\
Trust & .14** & -.51 [-.63,-.39]*** & -.27 [-.42,-.12]*** \\
\bottomrule
\end{tabular}}
\caption{Mediating Language Analysis: Analysis seeks to explain the correlation, $c = .36$, between \emph{percent Black} and \emph{residualized maternal mortality} through differences in language. $\alpha$: correlation between the theoretical factor and percent Black; $beta$: correlation between the theoretical factor and residualized maternal mortality. Reported Pearson r with 95\% confidence intervals in square brackets; ***$p <0.001$, **$p < 0.01$, *$p <.05$, after Benjamini-
--Hochberg correction. The $c-c'$ column uses a Sobel $p$ for significance~\cite{sobel1982asymptotic}.}
\label{table:mediation theoretical}
\end{table}

We considered each of the 4 theoretical dimensions as potential mediators. 
To zero in on explaining what is novel about the race-mortality, we controlled for all previously mentioned socioeconomic and risk factor variables by producing a residual of the variance left over. The correlation between percent Black and maternal mortality was then $c = .36$
Without this step, it could be that any mediators were simply accounting for socioeconomic or risk factor effects. As seen in Table \ref{table:mediation theoretical}, we found two of the theoretical dimensions and 3 of the topics had a significant mediation effect, in part explaining the disparity. 
For example, trust mediated the relationship -- the fact that communities expressing lower trust had greater maternal mortality, partially explained why Black percentage related to greater mortality.

\section{Discussion}\label{sec:discuss}

The results shown in this work demonstrate the efficacy of social media language to shed some light on community characteristics of maternal mortality. While social media data, by itself, is not able to reliably identify causes for high maternal mortality rates and disparities, it can provide supporting evidence for existing conjectures and generate hypotheses for further investigation. 

The observation that pregnancy-related topics, as well as the general 2,000 topics, both hold more predictive power than SES, risk factors, and race, combined, shows that such language-based data sets may contain characteristics of communities beyond that captured in standard variables used to study maternal mortality. Furthermore, the diversity of discussion themes in the pregnancy-related data set presents an opportunity to consider how different topics relate with maternal mortality rates and patterns of topic popularity across US counties. 

The novel mediation results presented in this work allow us to gain further insights into how \emph{affect, depression, stress}, and \emph{trust} relate to mortality rates and disparities. The results that \emph{trust} and \emph{affect} related significantly with mortality rates mirrors discussions from public health research: for instance, failure by hospitals, providers, and facilities to provide unbiased and nondiscriminatory care has already been shown to result in lower follow-up visits by Black and Latina women, which is believed to drive higher mortality rates. Trust in physicians and medical institutions has been extensively studied \cite{mechanic1996changing,hall2001trust,hall2002trust,hall2002measuring}, with multiple studies focusing on racial and ethnic differences in levels of trust~\cite{doescher2000racial,gordon2006racial,armstrong2007racial,armstrong2013prior}. Findings repeatedly show ethnic and racial differences in trust towards health-care systems, in addition to showing that distrust is associated with racial disparities in use of preventive services~\cite{musa2009trust}. The \emph{affect} result is also related to the \emph{Congratulatory Remarks} topic, indicating that communities with both more positive language and more positive discussions around pregnancy and birth may also be experiencing lower maternal mortality rates and disparities. These observations, along with existing discussions, provide potential actionable insights for policies at the community level. 

The results here are not without limitations: as with other studies heavily relying on social media data, there are inherent issues of selection bias in who is on the platform and which users meet the inclusion thresholds we set for the pregnancy-related and county-mapping data sets. There is also selection bias in tweets that are geo-located as well as language use by the individuals on Twitter compared to other platforms. It is imperative to not take these data sets as being representative of the U.S., the counties we study, or even individuals that maybe included in the data sets. 

Furthermore, we do not control for linguistic differences across different parts of the U.S. and some topics, as a result, may show significant spatial and geographic associations. Likewise, we set the seed-words for constructing the pregnancy-related data set using word2vec, which may also suffer bias issues: e.g, certain words which may be commonly used to discuss pregnancy and birth by certain groups of under-represented individuals may not pass this analysis. While we attempt to control for this by having a relatively large number of seed-words and instead relying on data cleaning, this remains a notable limitation. 

We were hindered by the availability of outcome data: a lot of the relevant data is available only at the county-level and crucial data like disparities by race were entirely unavailable. While we believe that studies like ours will provide additional data-sources, models, and measurements to further our understanding of maternal mortality and disparities, availability of ground truth data presents a significant bottleneck. The availability of ground truth data about mortality and disparities, including data regarding mortality rates for groups of individuals belonging to marginalized communities, as well as disaggregated data by different demographics such as race, age, education, income, and other demographics would allow for more fine-grained analysis. 
\section{Ethics Statement}
This study was reviewed by the University of Pennsylvania institutional review board and, due to lack of individual human subjects, found to be exempt. 
All data used in this study are publicly available. 
While the county-level language estimates are publicly available and will be posted on the project page\footnote{\label{projectpage}All data available at: \url{https://github.com/wwbp/maternal_mortality}}, the original tweets, which are also publicly available, are unable to be redistributed by the authors due to Twitter's Terms of Service. For additional privacy protection, we automatically replace any Twitter user names with <user> in our analysis and presentation in this paper.


\newpage

\newpage

\bibliographystyle{ACM-Reference-Format}
\bibliography{sample-base}


\newpage

\newpage

\appendix 
\section*{Appendix}
We include further details on results and discussions from the main text below: 

\section{Sample Tweets}

The last column shows a sample of three tweets for the topic. To find these representative tweets, we extract topic loadings over a random set of 500,000 pregnancy-related tweets. 
We then order the tweets by topic loadings and hand-select three tweets (out of the top ten) that best describe the topic, ignoring noisy or uninformative tweets. 
For example, a tweet ``teen rates!!!'' would load extremely high in our first topic, but it doesn't capture any additional information over the list of the highest-weighted words within the topic. Note that all typos and emoticons in the tweets are included unchanged.

\begin{table}[htpb!]\centering{\begin{tabular}{p{0.1\textwidth}p{0.345\textwidth}}\toprule
Topic Label & Sample Tweets \\ \hline
Teen \newline Pregnancy (1.34\%) &  teenage pregnancy \#iblamedavidcameron \newline
 Decreasing infant mortality around the world <URL> \newline 
 \#BecauseOfYolo teenage pregnancy rate has risen \\ \hline
Morning \newline Sickness (0.54\%) & The purpose of our lives is to give birth to the best which is within us~Marianne Williamson \#spirituality \newline
Ecotopic pregnancy diagnosis symptoms and complications <URL> \newline
Getting a sickness that isn't morning sickness while \#pregnant \#sucks \#cough \#throathurts \#stuffynose  \#blah \\ \hline
Celebrity \newline Pregnancies (1.42\%) &  Amber rose is pregnant ? \#damnwiz \newline
hopefully kim k's pregnancy doesnt last 72 days \newline
Taylor swift pregnant by harry \\ \hline
Abortion Rights (2.15\%) & Lawmakers ban shackling of pregnant inmates <URL> \newline
 \#SouthAfrica to care for all \#HIV positive infants <URL> \#worldaidsday \#womensrights \#children \newline
 Nebraska governor rejects prenatal care funding for illegal immigrants <URL> \\ \hline
Maternal \newline Studies (2.56\%) & Lower autism risk with folic acid supplements in pregnancy <URL> \newline
Postpartum cardiovascular risk linked to glucose intolerance during pregnancy <URL> \newline
Increased autism risk linked to hospital-diagnosed maternal infections <URL> \\ \hline
Congratulatory \newline Remarks (3.06\%) & Congrats to <USER> and <USER> on the birth of their baby \newline \#5yearsago i gave birth to my wonderful daughter <3 <3 <3 \newline Awwwwww my nephew's wife is pregnant <3 congrats! \\  \bottomrule
\end{tabular}}
\caption{Sample topics with sample tweets}
\label{table: example tweets}
\end{table}

\section{Theoretical Models}  

We present high-level details for each of the four models used in this paper. Detailed descriptions and evaluations can be found in the corresponding papers. Note that none of the models described below were developed for this paper.
\noindent \paragraph{Affect}
An affect model was built using a set of 2,895 annotated Facebook posts.  Each post was rated by two psychologists on a nine-point ordinal scale, based on the affective circumplex model introduced by \citet{russell1980circumplex}. A $\ell_2$ penalized linear (ridge) regression was built using 1---2grams extracted from each message. Using a 10-fold cross-validation setup, the ngram model resulted in a prediction accuracy (Pearson $r$) of 0.65. 
Full details can be found in \citet{preoctiuc2016modelling}.

\paragraph{Depression}
The MyPersonality data set~\cite{kosinski2015facebook}, which consisted of approximately 154,000 consenting users who shared Facebook statuses and completed a 100-item personality questionnaire was used. The personality questionnaire is based on the International Personality Item Pool proxy for the NEO Personality Inventory~\cite{costa1992normal}. 
This work then takes the average response to the seven depression-facet items (located within the larger Neuroticism scale) to estimate user-level degree of depression.
A ridge-penalized regression model was built~\cite{hoerl1970ridge} using a set of 2,000 LDA topics and 1-3grams extracted over 27,749 individuals and tested on 1,000 random individuals who used at least 1,000 words across all of their statuses. 
This resulted in a final prediction accuracy (Pearson $r$) of 0.39.
Full details can be found in \citet{schwartz2014towards}.

\paragraph{Trust}
Similar to the depression model, the trust model was built using the MyPersonality Facebook data set~\cite{kosinski2015facebook}. 
Consenting individuals were asked to share their Facebook statuses and answer a Big-Five personality questionnaire. The average of three of the ten trust-facet items from the agreeableness domain -- (1) "I believe that others have good intentions," (2)  "I trust what people say," and (3) "I suspect hidden motives in others" (reverse-coded) -- was used as a measure of trust. 
A predictive model was built on 26,243 users who answered the above question and also shared Facebook statuses (with at least 1,000 words across all statuses) and evaluated on a smaller set of users ($N=$621) who answered the ten trust-facet items. 
Using a set of 2,000 LDA topics and 1-3grams, this resulted in a prediction accuracy (Pearson $r$) of $0.35$.
Full details can be found in \citet{zamani2018predicting}.

\paragraph{Stress}
Participants were recruited through Qualtrics (an online survey platform, similar to Amazon Mechanical Turk), where each participant answered a series of demographic questions, the 10-item Cohen's Stress scale~\cite{cohen1997measuring} and consented to share their Facebook statuses. The analysis was then limited to those who self-reported age and gender (female/male) and who posted at least 500 words across all Facebook statuses, resulting in a final set of 2,749 participants.
A set of 2,000 Facebook topics were used as features in a ridge penalized regression model~\cite{hoerl1970ridge}. This resulted in a prediction accuracy of $0.32$ (Pearson $r$), using a 10-fold cross validation setup. Full details can be found in \citet{guntuku2019understanding}.

\section{Domain Transfer: Applying Facebook Models to Twitter Data}  

All four of our theoretical models were trained and evaluated on Facebook data in their original papers, whereas we applied the models to Twitter data. Some of the models have been shown to work in other domains (i.e., stress on Facebook vs Twitter; Guntuku 2019).  Additionally, previous work has found is that effect sizes tend to vanish without correcting for the domain transfer~\cite{rieman2017domain}, which we argue makes our prediction task harder. Additionally, \citet{rieman2017domain} showed that user-level Facebook models applied to county-level Twitter data are stable in terms of direction of effect sizes.

\section{Spatial Distributions}  

\begin{figure}[h]
\centering
\includegraphics[width=1\columnwidth]{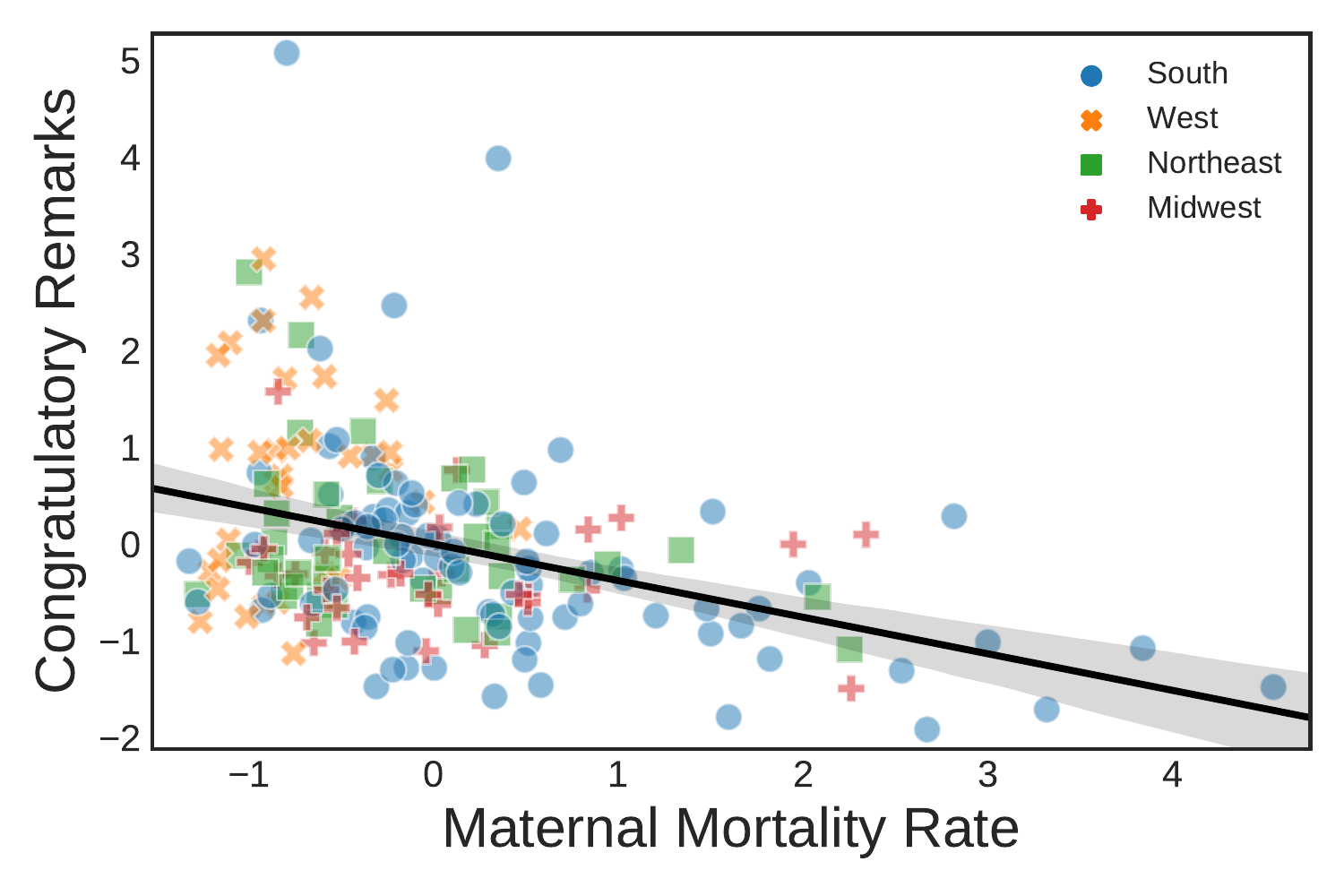}
\caption{Maternal mortality rate (residualized) vs the \emph{Congratulatory Remarks} topic loading. Dots are colored by which U.S. Census Region the county resides in: Midwest, Northeast, South and West.}
\label{fig:scatter topics}
\end{figure}

\begin{figure}[!b]
    \centering
    \begin{subfigure}[b]{.85\columnwidth}
        \includegraphics[width=\columnwidth]{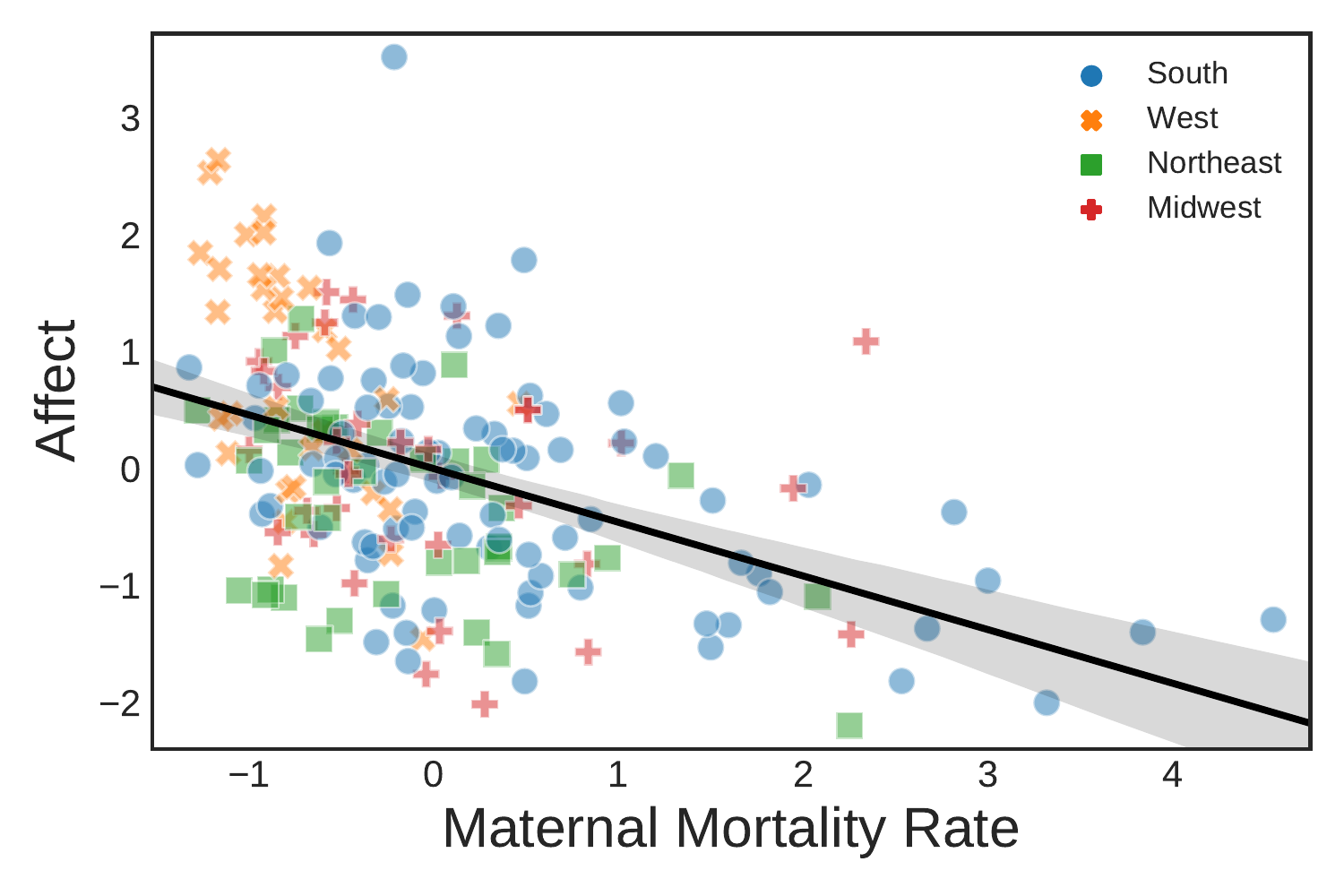}
        \caption{Affect}
        \label{fig:affect}
    \end{subfigure}
    ~ 
    
    \begin{subfigure}[b]{.85\columnwidth}
        \includegraphics[width=\columnwidth]{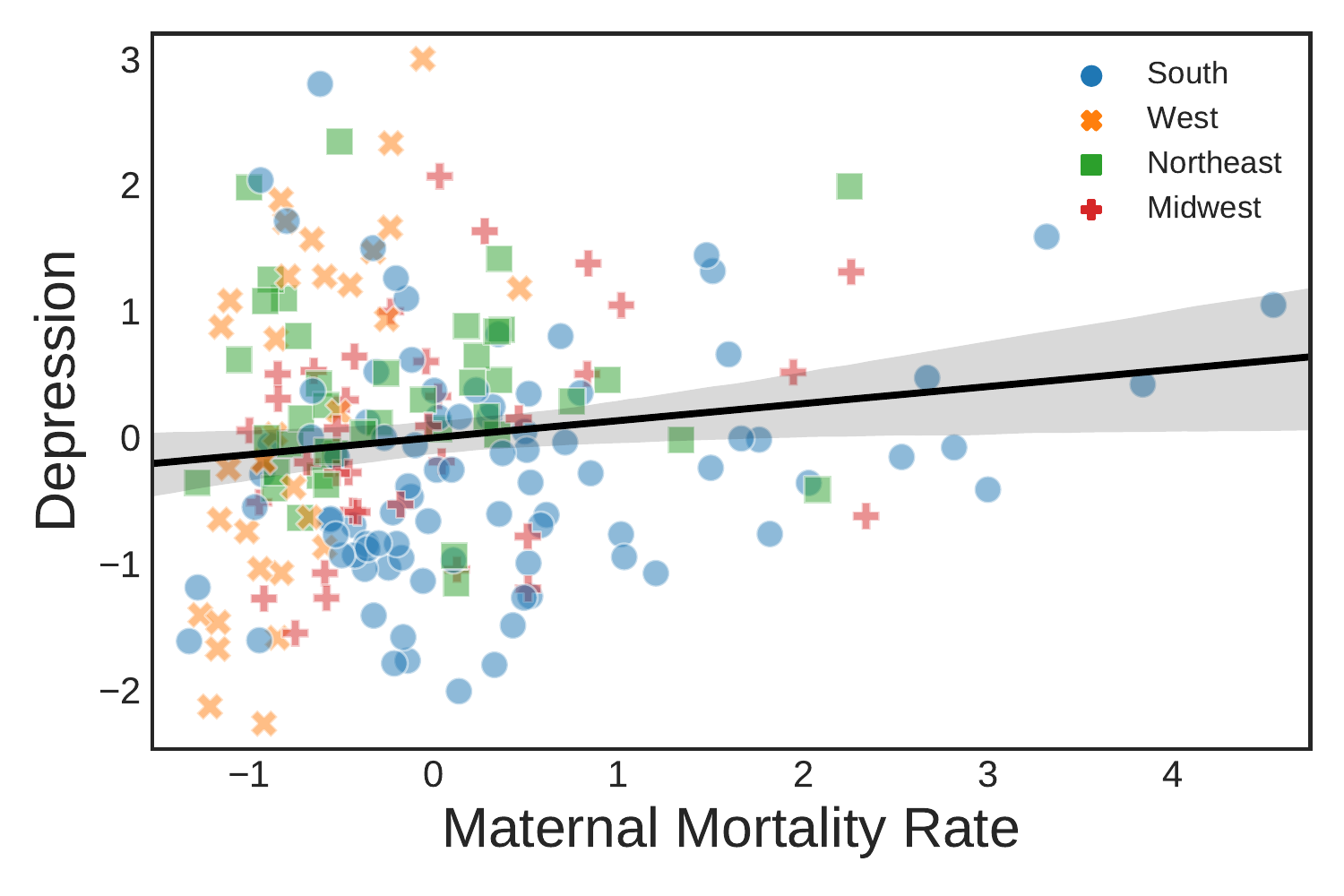}
        \caption{Depression}
        \label{fig:depression}
    \end{subfigure}
    ~ 

\begin{subfigure}[b]{.85\columnwidth}
        \includegraphics[width=\columnwidth]{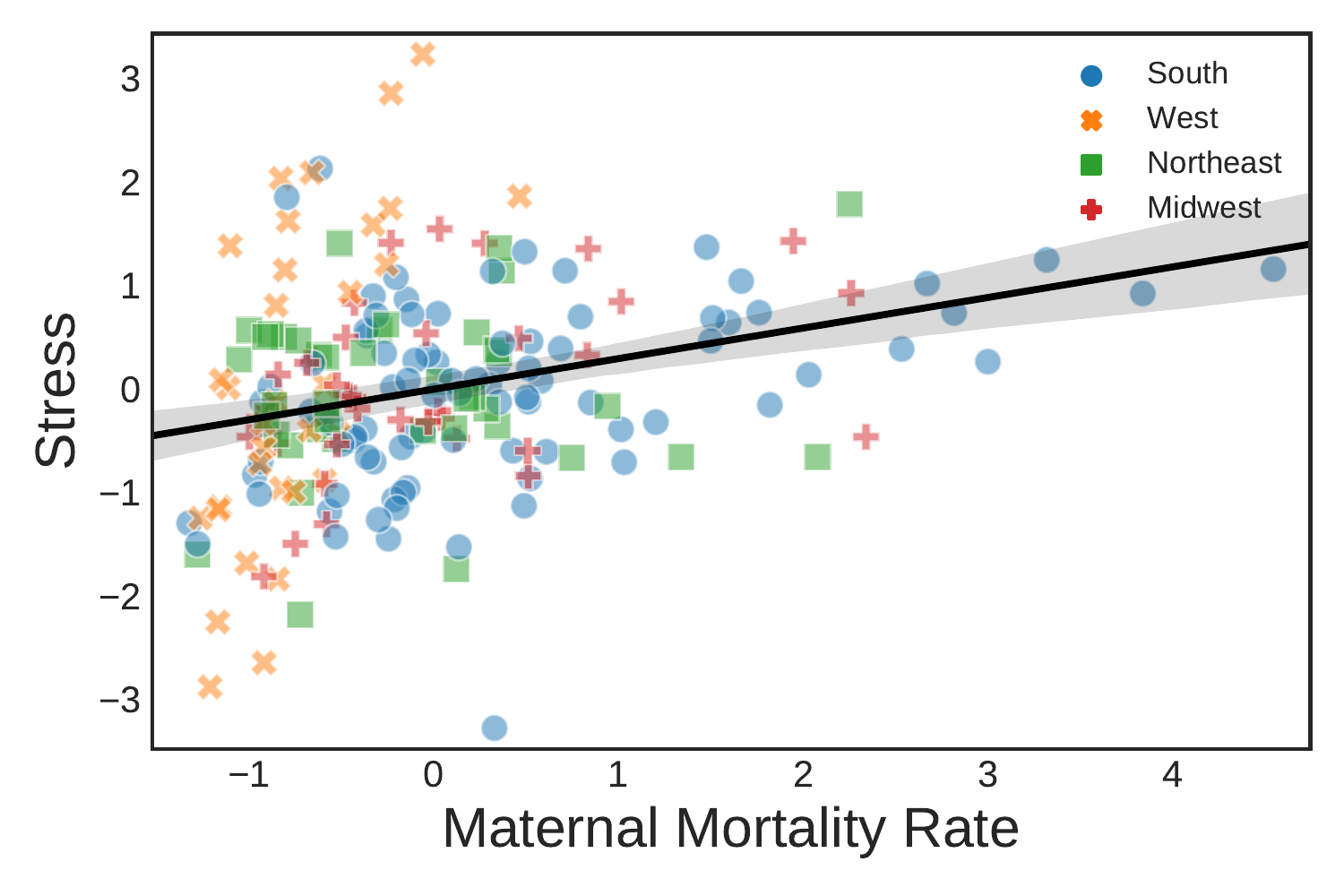}
        \caption{Stress}
        \label{fig:stress}
    \end{subfigure}
    ~ 
    
    \begin{subfigure}[b]{.85\columnwidth}
        \includegraphics[width=\columnwidth]{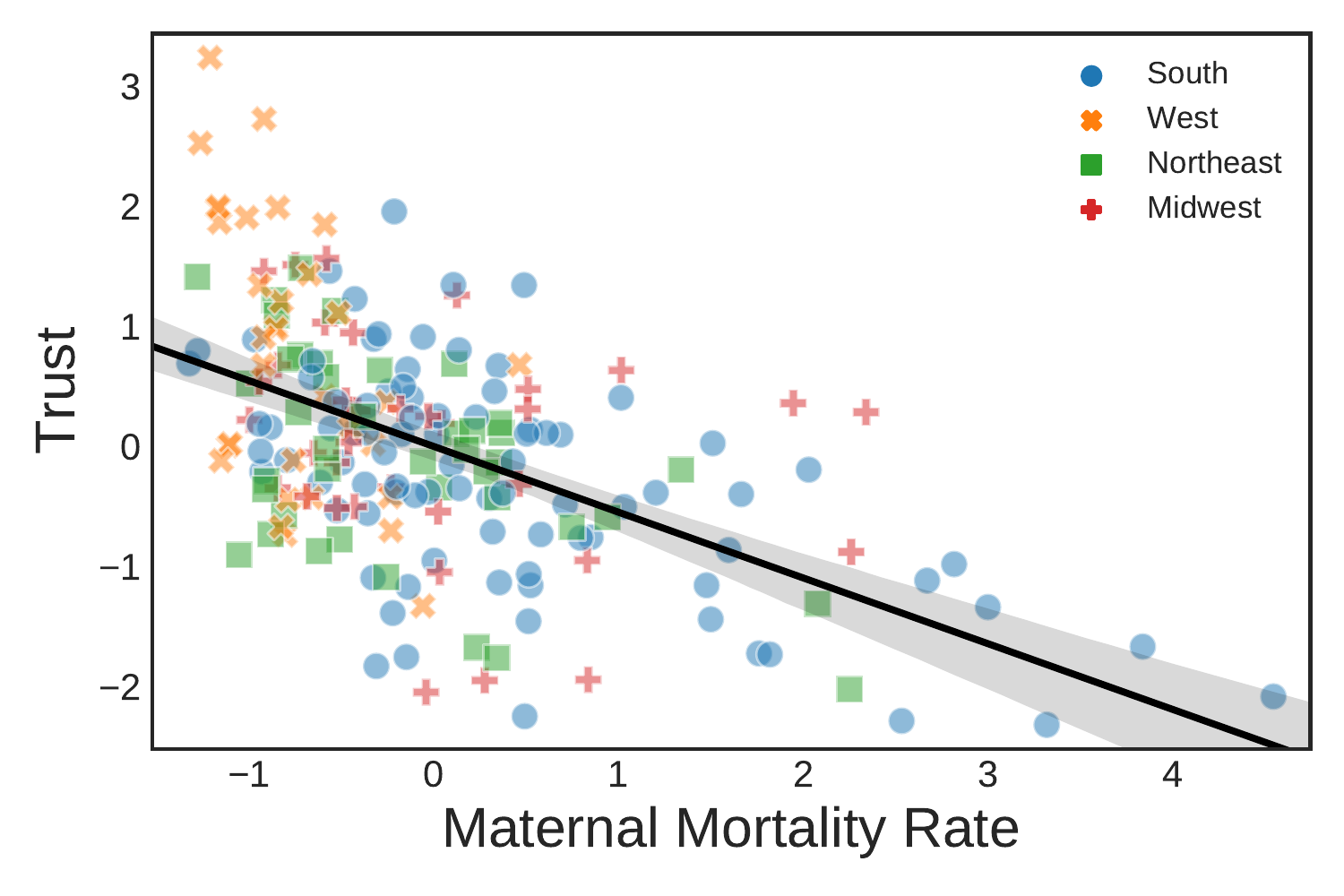}
        \caption{Trust}
        \label{fig:trust}
    \end{subfigure}
    \caption{Maternal mortality rate (residualized) vs theoretically relevant features. Dots are colored by which U.S. Census Region the county resides in: Midwest, Northeast, South and West.}\label{fig:theoretical scatter}
    \label{fig:scatter theoretical}
\end{figure}

Figure \ref{fig:scatter topics} shows the relationship between maternal mortality rates (residualized on race, median age of females, socioeconomics and risk factors) and the topic loadings for the \emph{Congratulatory Remarks} topic. Markers in the scatter plot are colored according to U.S. Census regions (Midwest, Northeast, South and West).  We see that lower usage of this topic is associated with high mortality rates. We also see spatial clustering across the regions. For example, the West tends to have lower rates of mortality but large variance in topic usage. The South has the most variation in mortality in addition to the largest outliers in topic usage. Figure \ref{fig:scatter theoretical} includes a similar set of plots for the theoretically-relevant features, showing significant associations between \emph{affect} and \emph{trust} and maternal mortality.

\end{document}